\documentclass{article}
\usepackage{times}
\usepackage{graphicx} 
\usepackage{subfigure} 

\usepackage{algorithm}
\usepackage{algorithmic}

\usepackage{natbib}

\usepackage{amsthm}
\usepackage{amsmath}
\usepackage{amsfonts}

\usepackage{algorithm}
\usepackage{algorithmic}

\newtheorem{thm}{Theorem}[section]
\newtheorem{cor}[thm]{Corollary}
\newtheorem{lem}[thm]{Lemma}

\theoremstyle{definition}

\newtheorem{prob}[thm]{Question}
\theoremstyle{remark}

\numberwithin{equation}{section}

\newcommand{\eps}{\varepsilon}

\def\sign{\operatorname{sign}}
\def\sortproc{\operatorname{SortProc}}
\def\quicksort{\operatorname{QuickSort}}
\def\onlinerank{\operatorname{OnlineRank}}
\def\alg{\operatorname{Alg}}
\def\E{{\mathbb E}}
\def\argmin{\operatorname{argmin}}

\def\R{\mathbb R}
\def\eps{\varepsilon}

\def\plackettluce{\operatorname{PlackettLuce}}

\def\Dfpl{D_{\operatorname{FPL}}}
\def\Afpl{A_{\operatorname{FPL}}}
\def\Rfpl{R_{\operatorname{FPL}}}

\newcommand{\beats}[3]{[{#1},{#2}]_{#3}}
\newcommand{\ind}[1]{{\bf{1}}_{#1}}
\newcommand{\pos}[1]{[#1]_+}
\newcommand{\forWarmuth}[1]{#1}

\begin{document}

\title{Online Ranking:  Discrete Choice, Spearman Correlation  and Other Feedback}
\author{Nir Ailon}




\def\sign{\operatorname{sgn}}
\def\dim{n}
\def\C{\mathbb C}
\def\P{\mathcal P}
\def\trace{\operatorname{tr}}
\def\diag{\operatorname{diag}}
\def\rank{\operatorname{rank}}
\def\F{{\mathcal F}}
\def\Id{\operatorname{Id}}
\def\poly{\operatorname{poly}}
\maketitle
\begin{abstract}
Given a set $V$ of $n$ objects, an online ranking system outputs at each time step a full ranking of the set, observes a feedback of some form and suffers a loss.  We study the setting in which the  (adversarial) feedback is an element in $V$, and the loss is the position ($0$th, $1$st, $2$nd...) of the item in the outputted ranking.  More generally, we study a setting in which the feedback is a subset $U$ of at most $k$  elements in $V$, and the loss is the sum of the positions of those elements.

We present an algorithm of expected regret $O(n^{3/2}\sqrt{Tk})$ over a time horizon of $T$ steps with respect to the best single ranking in hindsight.   This improves previous algorithms and analyses  either by a factor of either $\Omega(\sqrt{k})$, a factor of $\Omega(\sqrt{\log n})$ or by improving running time from quadratic to $O(n\log n)$ per round.   We also  prove a matching lower bound.  Our techniques also  imply an improved regret bound for online rank aggregation over the Spearman correlation measure, and to other more complex ranking loss functions.

\end{abstract}
\section{Introduction}
Many interactive online information systems (search, recommendation) present to a stream of users
 rankings of a set items in response to a specific query.  As feedback, these systems often observe a click (or a tap) on
one (or more) of these items.  Such systems are considered to be good  if users click on items that are
closer to the top of the retrieved ranked list, because it 
means they spent little time finding their sought information needs (making the simplifying assumption that  a typical user scans the list from
top to bottom).

We model this as the following iterative game.   There is a fixed set $V$ of $n$ objects. 
For simplicity, we first describe the \emph{single choice} setting in which for $t=1,\dots, T$, exactly one item $u_t$ from $V$ is chosen.
At each step $t$, 
the system outputs a (randomized) ranking $\pi_t$ of the set, and then $u_t$ is revealed to it.
The system loses nothing if $u_t$ is the first element in $\pi_t$, a unit cost if $u_t$ is in the second position,
$2$ units if it is in the third position, and so on.   The goal of the system is to minimize its total loss after $T$
steps.  (For simplicity we assume $T$ is known in this work.)    
The expected loss of the system is (additively) compared against that of the best (in hindsight) single ranking played throughout.   

More generally, nature can choose a subset $U_t\subseteq V$ per round.  We view
the set of chosen items in round $t$  as an indicator function  $s_t : V \mapsto \{0,1\}$ so that
$s_t(u)=1$ if and only if $u\in U_t$.
The loss function now penalizes the algorithm by the sum, over the elements of $U_t$, of
the positions of those elements in $\pi_t$.

We term such feedback as \emph{discrete choice}, thinking of the elements of $U_t$ as
items chosen by a user in an online system.
This paper studies online ranking over discrete choice problems, as well as over other more complex forms of feedback.
We derive both upper and lower regret bounds and improve on the state-of-the-art.

\subsection{Main Results}
For the discrete choice setting, we design an algorithm and derive bounds on its maximal expected regret as a function of $n, T$ and a 
uniform upper bound $k$ on $|U_t|$.
Our main result for discrete choice is given in Theorem~\ref{thm:main} below. Essentially, we show
an expected regret bound of  $O(n^{3/2}\sqrt{Tk})$ .  We argue in Theorem~\ref{thm:lower}
that this bound is  tight. The proofs of these theorems are given in Sections~\ref{sec:analysis} and~\ref{sec:analysis1}.
In Section~\ref{sec:compare} we compare our result to previous approaches.  To the best of our knowledge, our bound
is better than the best two previous approaches (which are incomparable): (1) We improve on Kalai et al.'s Follow the Perturbed Leader (FPL) algorithm's analysis \cite{Kalai:2005:EAO:1113185.1113189} by a factor of $\Omega(\sqrt k)$, and (2) We improve on  a more general algorithm by Helmbold et al. for learning permutations \cite{Helmbold:2009:LPE:1577069.1755841} by a factor of $\Omega(\log n)$.  It should be noted here, however, that  a more careful analysis of FPL results in 
regret bounds comparable with ours, and equivalently, a faster learning rate than that guaranteed in the paper \cite{Kalai:2005:EAO:1113185.1113189}.    (This argument will be explained in detail in Section~\ref{sec:FPL}.)

In Section~\ref{sec:ra}, we show that using our techniques, the problem of online rank aggregation over the Spearman
correlation measure, commonly used in nonparametric statistics  \cite{Spearman04},  also enjoys improved regret bounds.  This connects our work to 
 \cite{DBLP:journals/jmlr/YasutakeHTT12} on a similar problem with respect to the Kendall-$\tau$
distance.  

In the full version of this extended abstract we discuss a more general class of loss functions which assigns other importance weights to the various positions in the output ranking (other than the linear function defined above).  The result and the proof idea are presented in Section~\ref{sec:FPL}.

\subsection{Main Techniques}
Our algorithm maintains a weight vector $w \in \R^V$ which is updated at each step after nature reveals the subset $U_t$.
This weight vector is, in fact, a histogram counting the number of times each element appeared so far.
In the next round, it will use this weight vector as input to a noisy sorting procedure.\footnote{By this we mean, a procedure that outputs a randomized ranking of an input set.}
The main result in this work is, that as long as the noisy sorting procedure's output satisfies a certain property
 (see Lemma~\ref{lem:iia}), the algorithm has the desired
regret bounds.   Stated simply, this property ensures that for any fixed pair of items $u,v\in V$, the marginal distribution
of the order between the two elements follows a multiplicative weight update scheme with respect to $w(u)$ and $w(v)$.
We show that two noisy sorting procedures, one a  version of QuickSort
and the other based on a statistical model for rank data by Plackett and Luce, satisfy this property.
(We refer the reader to the book \cite{Marden95} for more details about the Plackett-Luce model in statistics.)



\section{Definitions and Problem Statement}\label{sec:def}
Let $V$ be a ground set of $n$ items.  A ranking $\pi$ over $V$ is an injection $\pi: V \mapsto [n]$, where $[n]$ denotes $\{1,2,\dots, n\}$.
We let $S(V)$ denote the space of rankings over $V$.
The expression $\pi(v)$ for $v\in V$ is the \emph{position} of $v$ in the ranking, where we think of lower positions as \emph{more favorable}.
For distinct $u,v\in V$, we say that $u \prec_\pi v$ if $\pi(u) < \pi(v)$ (in words: $u$ \emph{beats} $v$).  We use $\beats{u}{v}{\pi}$ as shorthand
for the indicator function of the predicate $u \prec_\pi v$.

At each step $t=1,\dots, T$ the algorithm outputs a ranking $\pi_t$ over $V$ and then
 observes a  subset $U_t\subseteq V$ which we also denote by its indicator function $s_t : V \mapsto \{0,1\}$.  The instantaneous loss  incurred by the algorithm at step $t$ is
\begin{equation}\label{eq:linloss}  \ell(\pi_t, s_t) = \pi_t\cdot s_t := \sum_{u\in V} \pi_t(u)s_t(u)\ ,\end{equation}
namely, the dot product of the $\pi_t$ and $s_t$, both viewed as vectors in $\R^n\equiv \R^V$.
Since in this work we are interested in bounding \emph{additive} regret, we can equivalently work with any loss function that
differs from $\ell$ by a constant that may depend on $s_t$ (but not on $\pi_t$).  This  work will take advantage of this
fact and will use the following \emph{pairwise} loss function, $\ell\ell$, defined as follows:

\begin{equation}\label{eq:lossdef} \ell\ell(\pi_t, s_t) := \sum_{u\neq v} \beats{u}{v}{\pi_t} \pos{s_t(v) - s_t(u)}\ ,\end{equation}
where $\pos{x}$ is $x$ if $x\geq 0$ and $0$ otherwise.
In words, this will introduce a cost of $1$ whenever $s_t(v)=1$, $s_t(u)=0$ and the pair $u,v$ is misordered in the sense that $u \prec_{\pi_t} v$.
A zero loss is incurred exactly if the algorithm places the elements in the preimage $s_t^{-1}(1)$ before the
elements in $s_t^{-1}(0)$. 
It should be clear that for any $s:V\mapsto \{0,1\}$ and $\pi\in S(V)$, the losses $\ell(\pi, s)$ and $\ell\ell(pi,s)$ differ
by a number that depends on $s$ only.
Slightly abusing notation, we define $$\ell\ell(\pi, s, u, v) :=  \beats{u}{v}{\pi} \pos{s(v) - s(u)} +  \beats{v}{u}{\pi} \pos{s(u) - s(v)}\ ,$$
so that $\ell\ell(\pi_t, s_t)$ takes the form $\sum_{\{u,v\}\subseteq V} \ell\ell(\pi, s, u, v)$.\footnote{Note that this expression makes sense because $\ell\ell(\pi, s, \cdot, \cdot)$ is symmetric in its last two arguments.}

Over a horizon of $T$ steps, the algorithm's total loss is $L_T(\alg) := \sum_{t=1}^T \ell\ell(\pi_t, s_t)$.  
We will compare the expected total loss of our algorithm with that of $\pi^* \in \argmin_{\pi \in S(V)}L_T(\pi)$, where $L_T(\pi) := \sum_{t=1}^T\ell\ell(\pi, s_t)$. \footnote{We slightly abuse notation by thinking of $\pi^*$ both as a ranking and
as an algorithm that outputs the same ranking at each step.}

Thinking of the aforementioned applications,  we say that $u$ is \emph{chosen} at step $t$ if and only if $s_t(u)=1$.
  In case exactly one item is chosen at each step $t$ we say that we are in the \emph{single choice}
setting.  If at most $k$ items are chosen we say that we are in the $k$-choice model.
Note that in the single choice case, the instantaneous losses $\ell$ and $\ell\ell$ at time each time $t$ are identical.

We will need an invariant  $M$ which measures a form of complexity of the value functions $s_t$, given as
\begin{equation}\label{eq:defM}
M =  \max_{t=1..T}\sum_{\{u,v\}} (s_t(v) - s_t(u))^2 \ .
\end{equation}
Note that since $s_t$ is a binary function, this is also equivalent to $M = \max_{t=1..T}\max_{\pi\in S(V)}\ell\ell(\pi, s_t)$,
namely,  the maximal loss of any ranking at any time step.  (Later in the discussion we will study  non-binary $s_t$, where this will not hold).
In fact, we need an upper bound on $M$, which (abusing notation) we will also denote by $M$.
In the most general case, $M$ can be taken as $n^2/4$ (achieved if exactly half of the elements are chosen).
In the single choice case, $M$ can be taken as $n$.  In the $k$-choice case, $M$ can be taken as $k(n-k) \leq nk$.
(We will assume always that $k\leq n/2$.)

\section{The Algorithm and its Guarantee for Discrete Choice}

Our algorithm $\onlinerank$  (Algorithm~\ref{alg:thealg}) takes as input the ground set $V$, a learning rate parameter $\eta \in [0,1]$, a reference to a randomized sorting procedure $\sortproc$ and
a time horizon $T$.    We present two possible randomized sorting procedures, $\quicksort$ (Algorithm~\ref{alg:quicksort}) and $\plackettluce$ (Algorithm~\ref{alg:plackettluce}).
 Both options satisfy an important property, described below in Lemma~\ref{lem:iia}.  Our main result for discrete choice is as follows.

\begin{thm}\label{thm:main}
Assume the time horizon $T$ is at least $n^2M^{-1}\log 2$.
If $\onlinerank$ is run with either $\sortproc=\quicksort$ or $\sortproc=\plackettluce$ and with $\eta = n\sqrt{\log 2}/\sqrt{TM}\leq 1$, then
\begin{equation}\label{eq:mainthm} \E[L_T(\onlinerank)] \leq L_T(\pi^*) +n\sqrt{TM\log 2}\ .\end{equation}
Additionally, the running time per step is $O(n\log n)$.
\end{thm}
The proof of the theorem is deferred to Section~\ref{sec:analysis}.  We present a useful corollary for the cases of interest.
\begin{cor}\label{cor:binary}
\begin{itemize}
\item 
In the general case,  if $T\geq 4\log 2$ and $\sortproc,\eta$ are as in Theorem~\ref{thm:main} (with $M=n^2/4$), then $\E[L_T(\onlinerank)] \leq L_T(\pi^*) +\frac {n^2} 2\sqrt{T\log 2}$.
\item 
In the $k$-choice case, if $T\geq nk^{-1}\log 2$ and $\sortproc, \eta$ are as in theorem~\ref{thm:main} (with $M=nk$), then
$\E[L_T(\onlinerank)] \leq L_T(\pi^*) + {n^{3/2}}\sqrt{ T k\log 2}$.
\item In the single choice casem if $T\geq n\log 2$  and $\sortproc, \eta$ are as in theorem~\ref{thm:main} (with $M=n$), then $\E[L_T(\onlinerank)] \leq L_T(\pi^*) + {n^{3/2}} \sqrt{ T\log 2}$.
\end{itemize}
\end{cor}
\noindent
We also have the following lower bound.
\begin{thm}\label{thm:lower}
There exists an integer $n_0$ and some function $h$ such that for all $n\geq n_0$ and $T\geq h(n)$,
for any algorithm, the minimax expected total regret in the single choice case after $T$ steps is at least $0.003 \cdot n^{3/2}\sqrt {Tk}$.
\end{thm}
Note that we did not make an effort to bound the function $h$ in the theorem, which relies on 
weak convergence properties guaranteed by the central limit theorem.  Better bounds could be derived by considering
tight convergence rates of binomial distributions to the normal distribution.  We leave this to future work.

\begin{algorithm}[t!] 
\caption{Algorithm $\onlinerank(V, \eta, \sortproc, T)$}
\begin{algorithmic}[1]
        \STATE given: ground set $V$, learning rate $\eta$, randomized sorting procedure $\sortproc$, time horizon $T$
	\STATE set $w_0(u) = 0$ for all $u\in V$
	\FOR {$t=1..T$}
		\STATE output $\pi_t = \sortproc(V, w_{t-1})$
		\STATE observe $s_t : V \mapsto \{0,1\}$
		\STATE set $w_t(u) = w_{t-1}(u) + \eta s_t(u)$ for all $u\in V$
	\ENDFOR
\end{algorithmic}
\label{alg:thealg}

\end{algorithm}

\begin{algorithm}[t!]
\caption{Algorithm $\quicksort(V, w)$}
\begin{algorithmic}[1]
	\STATE given: ground set $V$, score function $w: V\mapsto \R$
	\STATE choose $p\in V$ (pivot) uniformly at random
	\STATE set $V_L=V_R=\emptyset$
	\FOR {$v\in V$, $v\neq p$}
		\STATE with probability $\frac{e^{w(v)}}{e^{w(v)} + e^{w(p)}}$ add $v$ to $V_L$
		\STATE \ \ \ \  \ \ \ \ \ \ \ \ \ \ \ \ \ \ \ \ \ \  otherwise, add $v$ to $V_R$
	\ENDFOR
	\STATE return concatenation of  $\quicksort(V_L, w), p, \quicksort(V_R, w)$
\end{algorithmic}
\label{alg:quicksort}
\end{algorithm}

\begin{algorithm}[t!]
\caption{Algorithm $\plackettluce(V, w)$}
\begin{algorithmic}[1]
	\STATE given: ground set $V$, score function $w: V\mapsto \R$
	\STATE set $U=V$
	\STATE initialize $\pi(u) = \perp$ for all $u\in V$
	\FOR {$i=1..n(=|V|)$}
		\STATE choose random $u\in U$ with $\Pr[u] \propto e^{w(u)}$ \label{line:choose}
		\STATE set $\pi(u) = i$
		\STATE remove $u$ from $U$
	\ENDFOR
	\STATE return $\pi$
\end{algorithmic}
\label{alg:plackettluce}
\end{algorithm}

\section{Comparison With Previous Work}\label{sec:compare}

There has been much work on online ranking with various types of feedback and loss functions.
We are not aware of work that studies the exact setting here.

Yasutake et al. \cite{DBLP:journals/jmlr/YasutakeHTT12} consider online learning for \emph{rank aggregation}, where at
each step nature chooses a permutation $\sigma_t\in S(V)$, and the algorithm incurs the loss
$\sum_{u\neq v} \beats{u}{v}{\pi_t} \beats{v}{u}{\sigma_t}$.
Optimizing over this loss summed over $t=1,\dots, T$ is NP-Hard even in the offline setting \cite{DKNS}, while our problem, as we shall shortly see, is easy to solve offline.  Additionally, our problem is different and is not simply an easy instance of \cite{DBLP:journals/jmlr/YasutakeHTT12}.


A na\"ive, obvious approach to the problem of prediction rankings, which we state for the purpose of self containment, is by viewing each permutation
as one of $n!$ actions, and ``tracking'' the best permutation using a standard Multiplicative Weight (MW) update.  
Such schemes \cite{DBLP:conf/eurocolt/FreundS95,Littlestone:1994:WMA:184036.184040} guarantee an expected regret bound of
$O(M \sqrt{ Tn \log n})$.
The guarantee of Theorem~\ref{thm:main} is better by at least
a factor of $\Omega(\sqrt{n\log n})$ in the general  case, $\Omega(\sqrt{k\log n})$ in the $k$-choice case and $\Omega(\sqrt{\log n})$ in the single choice case.  
The distribution arising in the MW scheme would assign a probability proportional to $ \exp\{-\beta L_{t-1}(\pi)\}$ for
any ranking $\pi$ at time $t$, and for some learning rate $\beta>0$.  This distribution is not equivalent to neither $\quicksort$ nor $\plackettluce$, and it is not clear how to efficiently draw from it for large $n$.

\subsection{A Direct Online Linear Optimization View}

Our problem  easily lends itself to online linear optimization \cite{Kalai:2005:EAO:1113185.1113189}  over a discrete subset
of a real vector space.  In fact, there are multiple ways for doing this.

The loss $\ell$, as defined in Section~\ref{sec:def}, is a linear function of $\pi_t \in \R^n\equiv \R^V$.
The vector $\pi_t$ can take any vertex in the \emph{permutahedron}, equivelently, the set of vectors
with distinct coordinates over $\{0,\dots, n-1\}$.  It is easy to see that for any real vector $s$, minimizing $\pi\cdot s  = \sum \pi(u)s(u)$ is done by ordering the elements of $V$ in decreasing $s$-value $u_0,u_1,\dots, u_{n-1}$ and setting
$\pi(u_i)=i$ for all $i$.
The highly influencial paper of Kalai et al. \cite{Kalai:2005:EAO:1113185.1113189} suggests Follow the Perturbed
Leader (FPL) as a general approach for solving such online linear optimization problems.  The bound derived there
yields an expected regret bound of 
$O(n^{3/2}k\sqrt{T})$
for our problem.  This bound is comparable to ours for the single choice case,  is worse by a factor of $\Omega(\sqrt k)$ in the $k$-choice case and by a factor of $\Omega(\sqrt n)$ in the general case.
To see how the bound is derived, we remind the reader of how FPL works:  At time $t$, let $w_t(u)$ denote the number
of times $t'<t$ such that $u\in U_t$ (the number of appearances of $u$ in the current history).  The algorithm then outputs
the permutation ordering the elements of $V$ in decreasing $w_t(u)+\epsilon_u$ order, where for each $u\in V$, $\epsilon_u$ is an iid
real random variable uniformly drawn from an ``uncertainty'' distribution with a shape parameter that is controled by a chosen learning
rate, determined by the algorithm.  One version of FPL in \cite{Kalai:2005:EAO:1113185.1113189}, considers an uncertainty
distribution which is uniform in the interval $[0,1/\eta]$ for a shape parameter $\eta$.  The analysis there guarantees
an expected regret of 
$2\sqrt{\Dfpl\Afpl\Rfpl T}$
as long as $\eta$ is taken as 
$\eta = \sqrt{\frac{\Dfpl}{\Rfpl \Afpl T}}$,
   where $\Dfpl$  (here) is the diameter of the permutahedron in $\ell_1$ sense, $\Rfpl$ is defined as $\max_{t=1..n, \pi\in S(v)\subseteq\R^n} \pi\cdot s_t$ (the maximal per-step loss) and $\Afpl$ is the maximal
$\ell_1$ norm of the indicator vectors $s_t$.  A quick calculation shows that we have, for the $k$-choice case,  $\Dfpl = \Theta(n^2)$, $\Rfpl=\Theta(kn)$, $\Afpl=\Theta(k)$, giving the stated bound.

  As mentioned in the introduction, however, it seems that this suboptimal bound is due to the fact that
analysis of FPL should be done more carefully, taking advantage of the structure of rankings and of the loss functions we consider.  We further elaborate on this in Section~\ref{sec:FPL}.


Very recently, \cite{DBLP:conf/alt/SuehiroHKTN12} considered a similar problem, 
in a  setting in which the loss vector $s_t$ can be assumed to be anything with coordinates bounded by $1$.  
In particular, that result applies to the case in which $s_t$ is binary.  They obtain the same expected regret
bound, but with a per-step time complexity of $O(n^2)$, which is worse than our $O(n\log n)$. Their analysis takes
advantage of the fact that optimization over the permutahedron can be viewed as a prediction problem under submodular
constraints.

Continuing our comparison to previous results, Dani et al. \cite{DBLP:conf/nips/DaniHK07} provide for online linear oprimization problems a regret bound of
 \begin{equation}\label{eq:danietal} O( M\sqrt{T d \log d \log T})\ ,\end{equation} where $d$ is the ambient dimension
of the set $\{\pi\}_{\pi\in S(V)}\subseteq \R^n$.  Clearly $d=\Theta(n)$, hence this bound is worse than ours by
a factor of $\Omega(\sqrt{\log n \log T})$ in the single choice case and $\Omega(\sqrt{k\log n \log T})$
in the $k$-choice case.  

A less efficient embedding can be done in $\R^{n^2} \equiv \R^{V\times [n]}$ using the Birkhoff-vonNeumann embedding,
as follows.  Given $\pi \in S(V)$, we define the matrix $A_\pi\in \R^{n^2}$ by

$  A_\pi(u,i) = \begin{cases} 1 & \pi(u) = i \\ 0 & \mbox{otherwise} \end{cases}$.
For an indicator function $s : V\mapsto \{0,1\}$ we define the embedding $C_s\in \R^{n^2}$ by
$C_s(u,i) = \begin{cases} i & s(u) = 1  \\ 0 &\mbox{otherwise}\end{cases}$.
It is clear that $\ell'(\pi_t, s_t)$ defined above is equivalently given by $A_{\pi_t} \bullet C_{s_t} := \sum_{u,i} A_{\pi_t}(u,i) C_{s_t}(u,i)$.  Using the analysis of FPL \cite{Kalai:2005:EAO:1113185.1113189} gives an expected regret bound
of $O(n^2\sqrt{T})$ in the single choice case and $O(n^2 k\sqrt{T})$ in the $k$-choice case, which is worse
than our bounds by at least a factor of $\Omega(\sqrt {n})$ and $\Omega(\sqrt{nk)}$, respectively.

Another recent work that studied linear optimization over cost functions of the form $A_\pi \bullet C_t$ for general cost
matrices $C_t\in \R^{n^2}$ is that of Helmut and Warmuth \cite{Helmbold:2009:LPE:1577069.1755841}.  The expected regret
bound for that algorithm in our case is $O(n\sqrt{MT\log n} + n\log n)$ (assuming there is no prior upper bound on the total
optimal loss).\footnote{Note that one needs to carefully rescale the bounds to obtain a correct comparison with \cite{Helmbold:2009:LPE:1577069.1755841}.  Also, the variable $L_{EST}$ there, upper bounding the highest possible optimal loss, is computed by assuming all elements are chosen exactly $kT/n$ times.} This is worse by a factor of $\Omega(\sqrt{\log n})$ than our bounds.

\paragraph{Comparison of the Single Choice Case to Previous Algorithms for the Bandit Setting}\label{sec:comparesingle}
It is worth noting that in the single choice case, given $\pi_t$ and $\ell(\pi_t, s_t)$ it is possible to recover $s_t$ exactly.
This means that we can study the game in the single choice case in the so-called \emph{bandit setting}, where
the algorithm only observes the loss at each step.\footnote{Note that generally the bandit setting is more difficult
than the full-information setting, where the loss of all actions are known to the algorithm. The fact that the
two are equivalent in the single choice case is a special property of the problem.}
This allows us to compare our algorithm's regret guarantees to those of algorithms for online linear
optimization in the bandit setting.

Cesa-Bianchi and Lugosi  have studied the problem of optimizing  $\sum_{t=1}^T A_{\pi_t} \bullet C_t$
 in the 
bandit setting in  \cite{DBLP:journals/jcss/Cesa-BianchiL12}, 
where $A_{\pi_t}$ is the ranking embedding in $\R^{n^2}$ defined above.
They build on the methodolog of \cite{DBLP:conf/nips/DaniHK07}.
 They obtain an expected regret bound of $O(n^{2.5}\sqrt{T})$,  which is much worse
than the single choice bound in Corollary~\ref{cor:binary}.\footnote{This is not explicitly stated in their work, and requires plugging in  various calculations (which they provide) in the bound provided in their main theorem, in addition to scaling by $M= n$.}  Also, it is worth noting that the method for drawing a random ranking in each step in their
algorithm  relies on the idea of approximating the permanent, which is much more complicated than the algorithms 
presented in this work.

Finally, we mention the online linear optimization  approach in the bandit setting of Abernethy et al. \cite{DBLP:conf/colt/AbernethyHR08} in case the 
search is in a convex polytope.
The expected regret for our problem in the single choice setting using their approach  is $O(Md\sqrt{\theta(n) T})$,
where $d$ is the ambient dimension of the polytope, and $\theta(n)$ is a number that can be bounded by the number of its
facets \cite{private}. In the compact embedding (in $\R^n$), $d=n-1$ and $\theta(n)=2^n$.  In the embedding in 
$R^{n^2}$, we have $d=\Theta(n^2)$ and $\theta(n) = \Theta(n)$.  For both embeddings and for all cases we study, 
 the bound is worse than ours.

\paragraph{Comparison of Lower Bounds}
Our lower bound (Theorem~\ref{thm:lower}) is a refinement of  the lower bound in \cite{Helmbold:2009:LPE:1577069.1755841},
because the lower bound there was derived for a larger class of loss functions.  In fact, the method used there
for deriving the lower bound could not be used here.  
Briefly explained,  they reduce from simple online optimization over $n$ experts, each mapped to
a ranking so that no two rankings share the same element in the same position.  That technique cannot be used
to derive lower bounds in our settings, because all such rankings would have the exact same loss.

\section{Implications for Rank Aggregation}\label{sec:ra}

The (unnormalized) Spearman correlation  between two rankings $\pi, \sigma\in S(V)$, as
$ \rho(\pi, \sigma) = \sum_{u\in V}\pi(u)\cdot\sigma(u)$.

The corresponding \emph{online rank aggregation} problem, closely related to that of \cite{DBLP:journals/jmlr/YasutakeHTT12},
is defined as follows.  A sequence of rankings $\sigma_1,\dots, \sigma_T\in S(V)$ are chosen in advanced by the 
adversary.  At each time step, the algorithm outputs $\pi_t\in S(V)$, and then $\sigma_t$ is revealed to it.  The instantaneous loss
is defined as $-\rho(\pi_t, \sigma_t)$.  The total loss is $\sum_{t=1}^T(-\rho(\pi_t, \sigma_t))$, and the goal is to minimize the expected regret, defined with respect to $\min_{\pi\in S(V)} \sum_{t=1}^T (-\rho(\pi, \sigma_t))$.\footnote{For the purpose of rank aggregation, the Spearman correlation is something that we'd want to maximize.  We prefer to keep the mindset of \emph{loss minimization}, and hence work with $-\rho$ instead.}

Notice now that there was nothing in our analysis leading to Theorem~\ref{thm:main} that required $s_t$ to be a binary
function.  Indeed, if we identify $s_t\equiv -\sigma_t$, then the loss (\ref{eq:linloss}) is exactly $-\rho$.  Additionally,
the pairwise loss $\ell\ell$ (\ref{eq:lossdef}) satisfies that for all $\pi$ and $s_t\equiv \sigma_t$,
$ \ell(\pi, s_t)- \ell\ell(\pi, s_t) = C$,
where $C$ is a constant that depends on $n$ only.  To see why, one trivially verifies that when
moving from $\pi$ to a ranking $\pi'$ obtained from $\pi$  by swapping two \emph{consecutive} elements,
the two differences $\ell(\pi', s_t)-\ell(\pi, s_t)$  and $\ell\ell(\pi', s_t)-\ell\ell(\pi, s_t)$ are equal.  Hence again, we can consider
regret with respect to $\ell\ell$, instead of $\ell$.  The value of $M$ from (\ref{eq:defM}) is clearly $\Theta(n^4)$.
Hence, by an application of Theorem~\ref{thm:main}, we conclude the following bound for online rank aggregation over Spearman correlation:
\begin{cor}\label{cor:ra}
Assume a time horizon $T$ larger than some global constant.
If $\onlinerank$ is run  with either $\sortproc=\quicksort$ or $\sortproc=\plackettluce$, $s_t\equiv \sigma_t$ for $\sigma_t\in S(V)$ for all $t$
and $\eta = \Theta(1/(n\sqrt T))$, then the expected regret is at most $O(n^3\sqrt{T})$.
\end{cor}

A similar comparison  to  previous approaches can be done for the rank aggregation problem, as we did in Section~\ref{sec:compare}  for the cases of binary $s_t$.  Comparing with the direct analysis of FPL, the expected
regret would be $O(n^{3.5}\sqrt{T})$ (using $\Dfpl=\Theta(n^2), \Rfpl=\Theta(n^3), \Afpl=\Theta(n^2)$ here).
Comparing to \cite{Helmbold:2009:LPE:1577069.1755841}, we again  obtain here an improvement of $\Omega(\sqrt{\log n})$.

\section{Proof of Theorem~\ref{thm:main}}\label{sec:analysis}

Let $\pi^*$ denote an optimal ranking of $V$ in hindsight.
In order to analyze Algorithm~\ref{alg:thealg} with both $\sortproc=\quicksort$ and $\sortproc=\plackettluce$, we start with a simple lemma.
\begin{lem}\label{lem:iia}
The random ranking $\pi$ returned by $\sortproc(V, w)$ satisfies that for any given pair of distinct elements $u,v\in V$, the probability of the event $u \prec_\pi v$ equals
$ {e^{w(u)}}/({e^{w(u)} + e^{w(v)}})$, 
for both $\sortproc = \quicksort$ and $\sortproc = \plackettluce$.
\end{lem}
\noindent
The proof for case $\quicksort$ uses techniques from e.g. \cite{DBLP:journals/jacm/AilonCN08}.  
\begin{proof}
For the case $\sortproc = \quicksort$, the internal order between $u$ and $v$ can be determined in one of two ways.
(i) The element $u$ (resp. $v$) is chosen as pivot in some recursive call, in which $v$ (resp. $u$) is part of the input.
Denote this event $E_{\{u,v\}}$.
(ii) Some element $p\not\in \{u,v\}$ is chosen as pivot in a recursive call in which both $v$ and $u$ are part of the input, and in this recursive call the elements $u$ and $v$ are separated (one goes to the left recursion, the other to the right one).
Denote this event $E_{p; \{u,v\}}$.

It is clear that the collection of events $\{E_{\{u,v\}}\} \cup \{E_{p; \{u,v\}}: p\in V\setminus\{u,v\}\}$ is
a disjoint cover of the probability space of $\quicksort$.  If $\pi$ is the (random) output, then it is clear from the algorithm that
$$ \Pr[u \prec_\pi v | E_{\{u,v\}}] = e^{w(u)}/(e^{w(u)} + e^{w(v)})\ .$$
It is also clear, using Bayes rule, that for all $p\not\in\{u,v\}$,
\begin{eqnarray*} & &\Pr[u \prec_\pi v | E_{p; \{u,v\}}]  \\
&=& \frac{\frac{e^{w(u)}}{e^{w(u)}+e^{w(p)}}\frac{e^{w(p)}}{e^{w(p)}+e^{w(v)}}}{  \frac{e^{w(u)}}{e^{w(u)}+e^{w(p)}}\frac{e^{w(p)}}{e^{w(p)}+e^{w(v)}} + \frac{e^{w(v)}}{e^{w(v)}+e^{w(p)}}\frac{e^{w(p)}}{e^{w(p)}+e^{w(u)}}   } \\
&=&  e^{w(u)}/(e^{w(u)} + e^{w(v)}) \ , \\
\end{eqnarray*}
as required.
For the case $\sortproc = \plackettluce$, for any subset $X\subseteq V$ containing $u$ and $v$, let $F_X$ denote the event that, when the first of $u,v$ is chosen in Line~\ref{line:choose}, the value of $U$ (in the main loop) equals $X$.  It is clear that $\{F_X\}$
is a disjoint cover of the probability space of the algorithm.  If $\pi$ now denotes the output of $\plackettluce$, then the proof is completed by noticing that for any $X$,  
$ \Pr[u\prec_\pi v | F_X] = e^{w(u)}/(e^{w(u)} + e^{w(v)})$.
\end{proof}

The conclusion from the lemma is, as we show now, that  for each pair $\{u,v\}\subseteq V$ the algorithm plays a standard multiplicative update scheme
over the set of two possible actions, namely $u\prec v$ and $v \prec u$.  We now make this precise.
For each ordered pair $(u,v)$ of two distinct elements in $V$, let $\phi_t(u,v) = e^{-\eta\sum_{t'=1}^t \pos{s_{t'}(v) - s_{t'}(u)}}$.  
We also let $\phi_0(u,v)=1$.
On one hand, we have
\begin{eqnarray}\label{eq0}
 \sum_{\{u,v\}} &\log& \frac{\phi_T(u,v)+\phi_T(v,u)}{\phi_0(u,v) + \phi_0(v,u)}   \nonumber \\
&\geq& \sum_{u,v: u \prec_{\pi^*} v} \log {\phi_T(u,v)} - {n\choose 2}\log 2 \nonumber \\
&=& -\eta L_T(\pi^*) - {n\choose 2}\log 2\ .
\end{eqnarray}

\noindent
On the other hand,
\begin{eqnarray}
\sum_{\{u,v\}} &\log& \frac {\phi_T(u,v)+\phi_T(v,u)}{\phi_0(u,v) + \phi_0(v,u)} \nonumber \\
&=& \sum_{\{u,v\}}\sum_{t=1}^T \log \frac{\phi_t(u,v) + \phi_t(v,u)}{\phi_{t-1}(u,v) + \phi_{t-1}(v,u)}  \label{eq1} \nonumber \\
&=& \sum_{\{u,v\}}\sum_{t=1}^T \log \left (\frac{\phi_{t-1}(u,v)e^{-\eta \pos{s_t(v) - s_t(u)}}  }{\phi_{t-1}(u,v) + \phi_{t-1}(v,u)} \right . \nonumber \\
& &\ \ \ \ \ \ \ \ \ \ \ \ \ \ \ \ \ \ \  +\left . \frac{\phi_{t-1}(v,u)e^{-\eta \pos{s_t(u) - s_t(v)}}}{\phi_{t-1}(u,v) + \phi_{t-1}(v,u)} \right ) \nonumber
\end{eqnarray}
It is now easily verified that for any $u,v$,
\begin{eqnarray}\label{eq2}
& & \frac{\phi_{t-1}(u,v)}{\phi_{t-1}(u,v) + \phi_{t-1}(v,u)} \nonumber \\
 &=& \frac{1}{1 + e^{\eta\sum_{t'=1}^{t-1}\left( \pos{s_{t'}(v) - s_{t'}(u)} - \pos{s_{t'}(u) - s_{t'}(v)} \right)}} \nonumber \\
&=& \frac{1}{1 + e^{\eta \sum_{t'=1}^{t-1}\left ( s_{t'}(v) - s_{t'}(u)\right )}} = \frac {1}{1 + e^{w_{t-1}(v) - w_{t-1}(u)}} \nonumber \\
&=& \frac{e^{w_{t-1}(u)}}{e^{w_{t-1}(u)} + e^{w_{t-1}(v)}}\ . 
\end{eqnarray}

Plugging (\ref{eq2}) in (\ref{eq1}) and using Lemma~\ref{lem:iia}, we conclude
\begin{eqnarray}\label{eq1a}
\sum_{\{u,v\}} &\log& \frac {\phi_T(u,v)+\phi_T(v,u)}{\phi_0(u,v) + \phi_0(v,u)}  \\
&=& \sum_{\{u,,v\}} \sum_{t=1}^T \log \E \left [ e^{-\eta \ell(\pi_t, s_t, u, v)}\right ]  \nonumber \\
&\leq& \sum_{\{u,v\}} \sum_{t=1}^T \left ( -\E\left [ \eta \ell(\pi_t, s_t, u, v)\right ] \right . \nonumber \\ 
& & \ \ \ \ \ \ \ \ \ \ \ \ \ \ \ \ \ \ \ \ \left . +  \E\left [ \eta^2 \ell^2(\pi_t, s_t, u, v )/2  \right]\right ) \nonumber \\
&\leq& \sum_{\{u,v\}} \sum_{t=1}^T -\eta \E\left [  \ell(\pi_t, s_t, u, v)\right ] + \eta^2 TM/2 \nonumber \\
&=&  -\eta\E[L_T] + \eta^2TM/2\ ,
\end{eqnarray}
where we used the fact that $e^{-x} \leq 1-x+x^2/2$ for all $0 \leq x \leq 1$, and that $\log (1+x) \leq x$ for all $x$.
Combining (\ref{eq1a}) with (\ref{eq0}), we get
\begin{equation*}
\E[L_T] \leq \eta TM/2 +  L_T(\pi^*) + \eta^{-1}{n\choose 2} \log 2 \ .
\end{equation*}
Setting $\eta = n\sqrt {\log 2} /\sqrt{TM}$, we conclude the required.

\section{Proof of Theorem~\ref{thm:lower}}\label{sec:analysis1}

We provide a proof for the single choice case in this extended abstract, and include notes fo the $k$-choice case within the proof.
For the single choice case, recall that the losses $\ell$ and $\ell\ell$ are identical.

Fix $n$ and $V$ of size $n$, and assume $T \geq 2n$.
Assume the adversary chooses the sequence $u_1,\dots, u_T$ of single elements so that each element $u_i$ is chosen
independently and uniformly at random from $V$. [For general $k$, we will select subsets $U_1,\subset,U_T$ of size
$k$ at each step, uniformly at random from the space of such subsets].
For each $u\in V$, let $f(u)$ denote the frequency of $u$ in the sequence, namely
$f(u) = |\{i: u_i=u\}|$.
Clearly, the minimizer $\pi^*$ of $L_T(\pi)$ can be taken to be any ranking $\pi$ satisfying
$ f({{\pi}^{-1}(1)}) \geq f({{\pi}^{-1}(2)})\geq \cdots \geq f({{\pi}^{-1}(n)})$.
For ease of notation we let $u^j = {{\pi^*}^{-1}(j)}$, namely the element in position $j$ in $\pi^*$.
The cost $L_T(\pi^*)$ is given by
$L_T(\pi^*) = \sum_{j=1|}^n f({u^j}) (j-1)$.
For any number $x\in [0,T]$, let $m(x) = |\{u\in V: f(u) \geq x\}|$, namely, the number of elements with
frequency at least $x$.
Changing order of summation, $L_T(\pi^*)$ can also be written as

$L_T(\pi^*) = \sum_{x=1}^T (0 + 1 + 2 + \cdots + (m(x)-1))=\frac 1 2\sum_{x=1}^T m(x)(m(x)-1)$.
This, in turn, equals $\frac 1 2 \sum_{x=1}^T \sum_{u\neq v} \ind{f(u) \geq x} \ind{f(v)\geq x}$.

By linearity of expectation, $\E[L_T(\pi^*)]  = \frac 1 2 \sum_{x=1}^T \sum_{u\neq v} \E[\ind{f(u) \geq x} \ind{f(v)\geq x}]$.  This clearly equals $\frac 1 2 n(n-1) \sum_{x=1}^T \E[\ind{f(u^*) \geq x} \ind{f(v^*)\geq x}]$,
where $u^*, v^*$ are any two fixed, distinct elements of $V$.
Note that $f(u)$ is distributed $B(T,1/n)$ for any $u\in V$,
where $B(N,p)$ denotes Binomial with $N$ trials and probability $p$ of success. 
In what follows we let $X_{N,p}$
be a random variable distributed $B(N,p)$.
Let $\mu= T/n$ by the expectation of $X_{T,1/n}$, and let $\sigma = \sqrt{T(n-1)}/n$ be its standard deviation.
[For general $k$, instead, we have moments of a the binomial with $n$ trials and probability $k/n$ of success.]
We will assume for simplicity that $\mu$ is an integer (although this requirement can be easily removed).
We will fix an integer $j>0$ that will be chosen later.
We split the last expression as $\E[L_T(\pi^*)] = \alpha+\beta+\gamma$, where
\begin{eqnarray*} 
\alpha&=& \frac 1 2 n(n-1) \sum_{x=1}^{\mu - \lfloor j\sigma\rfloor -1} \E[\ind{f(u^*) \geq x} \ind{f(v^*)\geq x}] \\
\beta&=& \frac 1 2 n(n-1) \sum_{x= \mu - \lfloor j\sigma \rfloor}^{\mu+\lfloor j\sigma\rfloor} \E[\ind{f(u^*) \geq x} \ind{f(v^*)\geq x}] \\
\gamma&=& \frac 1 2 n(n-1) \sum_{x= \mu + \lfloor j\sigma\rfloor + 1}^{T} \E[\ind{f(u^*) \geq x} \ind{f(v^*)\geq x}] \ .\\
\end{eqnarray*}

Before we bound $\alpha, \beta,\gamma$, first note that for any $x$, the random variable $(f(u^*) | f(v^*) = x)$ is distributed
$B(T-x, 1/(n-1))$.  Also, for any $x$ the function $g(x') = \Pr[f(u^*) \geq x | f(v^*) = x']$ is monotonically decreasing in $x'$.
  Hence, for any $1\leq x\leq T$,
\begin{eqnarray}
&E&[\ind{f(u^*)\geq x}\ind{f(v^*)\geq x}] \\ \nonumber \\
&=& \sum_{x'=x}^T \Pr[f(v^*)= x'] \cdot\Pr[f(u^*) \geq x | f(v^*) = x'] \nonumber \\
&\leq&  \sum_{x'=x}^T \Pr[f(v^*)= x'] \cdot \Pr[f(u^*) \geq x | f(v^*) = x] \nonumber \\
&=& \Pr[f(v^*)\geq x] \cdot \Pr[f(u^*) \geq x | f(v^*) = x] \nonumber \\
&=& \Pr[X_{T, 1/n} \geq x] \cdot \Pr[X_{T-x, 1/(n-1)} \geq x]  \label{eq:abc}
\end{eqnarray}

\paragraph{Bounding $\gamma$:} We use Chernoff bound, stating that for any integer $N$ and probability $p$, 
\begin{eqnarray}
\forall x\in[Np, 2Np],\ \ \ \ \ \ \ \  & & \nonumber \\
\Pr[X_{N,p} \geq x] &\leq& \exp\left \{ \frac{- (x-Np)^2}{(3Np) }\right \}\ . \label{eq:cher1} \\
\forall x>2Np,\ \ \ \ \ \ \ \ & & \nonumber \\
 \Pr[X_{N,p} \geq x] &\leq& \Pr[X_{N,p} \geq 2NP] \ . \label{eq:cher2}
\end{eqnarray}

Plugging (\ref{eq:abc}) in the definition of $\gamma$ and using (\ref{eq:cher1}-\ref{eq:cher2}), we conclude that there
exists global integers $j, n_0$ and a polynomial $P$ such that for all $n \geq n_0$ and $T\geq P(n)$, 
\begin{equation}\label{eq:boundgamma}
\gamma   \leq  0.001\cdot  n(n-1) \sqrt{T/n} \leq   0.001 \cdot n^{3/2}\sqrt T\ .
\end{equation}

\paragraph{Bounding $\beta$:}
Using the same $j$ as just chosen, possibly increasing $n_0$ and applying the central limit theorem, we conclude that
there exists a function $h$ such that for all $n\geq n_0$ and $T\geq h(n)$,

\begin{equation}\label{eq:beta}
\beta \leq \frac 1 2 n(n-1)\left(\sqrt{\frac T n}+1\right )\sum_{i=-j}^j (1-\Phi(i - 1/100))^2\ ,
\end{equation}
where $\Phi$ is the normal cdf.   For notation purposes, let $\Psi(x) = 1-\Phi(x)$ and $\epsilon = 1/100$.  Hence,
\begin{eqnarray*}\label{eq:beta1}
\beta &\leq& \frac 1 2 n(n-1)\left(\sqrt{\frac T n}+1\right )  \\
& &\ \ \ \ \times \left (\Phi(-\epsilon)^2  + \sum_{i=1}^j \left ( \Phi(i-\epsilon)^2 + \Psi(i+\epsilon)^2  \right ) \right  )\ .\\
\end{eqnarray*}
We now make some rough estimates of the normal cdf.  The reason for doing these tedious calculations will be 
made clear shortly. One verifies that  $\Phi(-\epsilon) \leq 0.497$, $\Phi(1-\epsilon)\leq 0.839$, $\Phi(2-\epsilon)\leq 0.977$, $\Phi(3-\epsilon)\leq 0.999$, $\Psi(1+\epsilon) \leq 0.157$, $\Psi(2+\epsilon)\leq 0.023$, $\Psi(3+\epsilon)\leq 0.001$.
Hence,

\begin{eqnarray*}\label{eq:beta2}
\beta &\leq& \frac 1 2 n(n-1)\left(\sqrt{\frac T n}+1\right )   \\
& & \ \ \ \ \times  \left ( 2.929 +  \sum_{i=4}^j \left ( \Phi(i-\epsilon)^2 + \Psi(i+\epsilon)^2  \right )  \right )\\
\end{eqnarray*}
It is now easy to verify using standard analysis that for all $i\geq 4$, 
\begin{equation}\label{eq:cdf} \Phi(i-\epsilon)^2 + \Psi(i+\epsilon)^2  \leq 1\ . \end{equation}
\noindent
Therefore,
\begin{eqnarray*}\label{eq:beta3}
\beta &\leq& \frac 1 2 n(n-1)\left(\sqrt{\frac T n}+1\right ) (j - 0.07) \\
&\leq& \frac 1 2 n^{3/2}\sqrt T(j-0.07) + \frac 1 2 n^2(j-0.07)
\end{eqnarray*}
(Note that the crux of the  enitre proof is in getting the first summand in the last expression to be $\frac 1 2 n^{3/2}\sqrt T (j-c)$ for some $c>0$ .  This is the  reason we needed to estimate the normal cdf around small integers, and the inequality (\ref{eq:cdf}) for larger integers.)
\paragraph{Bounding $\alpha$} is done trivially by using  $ \E[\ind{f(u^*) \geq x} \ind{f(v^*)\geq x}]\leq 1$.
This gives,
\begin{eqnarray*}\label{eq:alpha}
\alpha &\leq& \frac 1 2 n(n-1)\left (\mu - \lfloor j\sigma \rfloor-1\right )  \\
&\leq & \frac 1 2 n(n-1)\left ( T/n - j\sqrt{T/n}+1 \right ) \\
&\leq& \frac 1 2 (n-1) T - \frac 1 2j  n^{3/2}\sqrt T  + \frac 1 2 j \sqrt{Tn}  + \frac 1 2 n^2
\end{eqnarray*}

\paragraph{Combining our bound for} $\alpha, \beta, \gamma$, possibly increasing $n_0$ and the function $h$, we conclude that there exists a global integer  $n_0$ and a function $h$ such that for all $n\geq n_0$ and $T \geq h(n)$,
$$ \E[L_T(\pi^*)]  = \alpha + \beta + \gamma \leq \frac 1 2 (n-1)T - 0.003\cdot n^{3/2} \sqrt T\ .$$

On the other hand, we know that for any algorithm, the expected total loss is exactly $\frac 1 2 T(n-1)$.
Indeed, each element $u_t$ in the sequence $u_1,\dots, u_T$ can be assumed to be  randomly drawn after $\pi_t$
is chosen by the algorithm, hence, the expected loss at time $t$ is exactly $(0+1+\cdots+(n-1))/n = (n-1)/2$.
This concludes the proof.

\forWarmuth{
\section{PlackettLuce, FPL, More Interesting Loss Functions and Future Work}\label{sec:FPL}

Our main algorithm OnlineRank (Algorithm~\ref{alg:thealg})  with the PlackettLuce  procedure (Algorithm~\ref{alg:plackettluce}) is, in fact, an FPL implementation
with the uncertainty distribution  chosen to be \emph{extreme value of type 1}.\footnote{A also often known as the \emph{Gumble} distribution.}  This distribution has a cdf of $F(x) = e^{-e^{-x}}$.
A proof of this fact cat be found in \cite{Yellott77}.  
We chose a different analysis because (noisy) QuickSort is an important and interesting algorithm to analyze, while it is \emph{not} equivalent to FPL.

The basic idea of our analysis in Section~\ref{sec:analysis} was, in view of the pairwise decomposable loss $\ell\ell$,  to show that we could accordingly execute a multiplicative
weights  algorithm simultaneously for each pair of elements, over a binary set of actions consisting of the
two possible ways of ordering the pair.   Any FPL scheme (in $\R^2$, not in $\R^n$!) could have been used to replace the multiplicate weight update.
The key was to notice that, at each time step $t$, at most $nk$ pairs could contribute to the loss, while the remaining pairs
contribute nothing, regardless of the action chosen for them.

Consider now a more general setting, in which our loss function is defined as $\ell_z(\pi_t, s_t) = \sum_{u\in U} z(\pi_t(u))\cdot s_t(u)$, where the parameter $z : [n]\mapsto \R$ is a monotone nondecreasing weight function, assigning different importance to the $n$ possible positions.  We studied the linear function $z=z_{\operatorname{LIN}}$ with $z_{\operatorname{LIN}}(i)=i-1$.  Other important functions are, for example $z_{\operatorname{NDCG}}$ defined as $z_{\operatorname{NDCG}}(i)=1/log_2(i+1)$, related
to the commonly used NDCG measure from information retrieval \cite{Jarvelin:2002:CGE:582415.582418}.  In the full version,
we will prove the following result:
\begin{thm}\label{eq:forfull}
Assume $z(i) = \alpha_0 + \sum_{j=1}^d \alpha_j i^j$ for some constant degree $d\geq 1$ and constants $\alpha_1,\dots, \alpha_d\geq 0$, with $\alpha_d>0$.
Also assume that  $s_t$ is a $k$-choice indicator function.
Then it is possible to set the shape parameter $\varepsilon$ of FPL \cite{Kalai:2005:EAO:1113185.1113189} so that
the expected online regret of its output 
is $O(n^{d+1/2}\sqrt{Tk})$, with respect to the best ranking in hindsight.
\end{thm}
Note that the analysis of \cite{Kalai:2005:EAO:1113185.1113189} results in bounds that are  worse by a factor of $\Omega(\sqrt{k})$ (which can be as high as $\Omega(\sqrt n)$),
and the bounds of \cite{Helmbold:2009:LPE:1577069.1755841} are worse by a factor of $\Omega(\sqrt{\log n})$.

The analysis, which will appear in the full version, relies on the ability to decompose the instantaneous loss $\ell_z$
over all subsets of $V$ of sizes $2$, $3$, \dots, $(d+1)$.

Theorem~\ref{eq:forfull} does not apply to functions such as $z_{\operatorname{NDCG}}$, which leaves open the following.

\begin{prob}
What are the correct minimax regret bounds over a given loss function $\ell_z$   for a given monotone nondecreasing $z$, 
and feedback $s_1..s_T$ from a given family of functions, as $n$ grows?
Is it always better by a factor of $\Omega(\sqrt{\log n})$ than the bound in \cite{Helmbold:2009:LPE:1577069.1755841}?}
\end{prob}

Another major open question is the following.
We argued in Section~\ref{sec:compare}, that the single choice case is also equivalently a bandit setting, because if we
only observe $\ell(\pi_t, s_t)$ then we can recover $s_t$.  This however is obviously not the case for the $k$-choice setting for $k>1$. 
\begin{prob}
What can be done in the bandit setting?  Is the algorithm CombBand of \cite{DBLP:journals/jcss/Cesa-BianchiL12}
the optimal for the setting studied here?
\end{prob}

\bibliography{online_AUC}
\bibliographystyle{icml2014}
\end{document}